\pdfoutput=1

\documentclass[11pt]{article}

\usepackage[final]{acl}

\usepackage{times}
\usepackage{latexsym}
\usepackage{amsmath}
\usepackage{amssymb}
\usepackage{bm}

\usepackage[T1]{fontenc}

\usepackage[utf8]{inputenc}

\usepackage{microtype}

\usepackage{inconsolata}
\usepackage{tcolorbox}
\usepackage{booktabs}

\usepackage{graphicx}
\usepackage{subcaption}
\usepackage{caption}

\title{Mind the Gap: Entity-Preserved Context-Aware ASR Structured Transcriptions}

\author{Duygu Altinok \\
  Independent Researcher, Germany \\
  \texttt{duygu.altinok@onlyduygu.com}}

\begin{document}
\maketitle
\begin{abstract}
Automatic Speech Recognition (ASR) systems, such as Whisper, achieve high transcription accuracy but struggle with named entities and numerical data, especially when proper formatting is required. These issues increase word error rate (WER) and impair semantic understanding in critical domains like legal, financial, and medical applications. We propose a novel training approach that extends the semantic context of ASR models by adding overlapping context windows during training. By sliding 5-second overlaps on both sides of 30-second chunks, we create a 40-second "effective semantic window," improving entity recognition and formatting while focusing predictions on the central 30 seconds. To address entities spanning chunk boundaries, we reassign such entities entirely to the right-hand chunk, ensuring proper formatting. Additionally, enriched training data with embedded entity labels enables the model to learn both recognition and type-specific formatting. Evaluated on the Spoken Wikipedia dataset, our method improves performance across semantic tasks, including named entity recognition (NER) and entity formatting. These results highlight the effectiveness of context-aware training in addressing ASR limitations for long-form transcription and complex entity recognition tasks.
\end{abstract}
\section{Introduction}
Recent advancements in speech and natural language processing (NLP) have significantly improved the capabilities of automatic speech recognition (ASR) systems. Large-scale models like Whisper \cite{radford2022robustspeechrecognitionlargescale} have achieved remarkable transcription accuracy. However, challenges remain in handling named entities and numerical data. ASR models often fail with entities like names, organizations, and numerical data (e.g., phone numbers, percentages, or monetary values), particularly when proper formatting is required. These shortcomings not only increase word error rate (WER) but also degrade semantic understanding, which is vital in domains such as legal, financial, and medical applications.

Named Entity Recognition (NER) from speech is a key task in spoken language understanding, where entities such as person names (PERSON), geographical locations (GPE), and organizations (ORG) are identified and classified. Traditional pipeline systems \cite{Jannet2017InvestigatingTE} employ an ASR module followed by a text-based NER model, but these systems suffer from error propagation due to transcription inaccuracies. In contrast, end-to-end (E2E) systems \cite{Gaido_2023,serdyuk2018endtoendspokenlanguageunderstanding,yadav2020endtoendnamedentityrecognition,mdhaffar2022endtoendmodelnamedentity,arora2022tokenlevelsequencelabelingspoken} directly extract entities from speech, bypassing intermediate ASR outputs and mitigating error propagation. However, ASR-generated output often lacks capitalization, punctuation, and other indicators necessary for accurate NER, making the task inherently more difficult.

An additional challenge is entity formatting. For example, "two one two five three two one" could be formatted as "212-5321" if representing a phone number but as "2125321" for a student ID. Proper formatting requires recognizing the entity type and applying the correct format. This complexity is further exacerbated when entities span chunk boundaries during inference, which can lead to errors. For instance, a numerical entity split across two chunks may be misinterpreted as two separate entities, resulting in incorrect formatting and increased WER.

In this work, we address these limitations by introducing a novel training approach for ASR systems. We propose extending the semantic window during training to improve entity recognition and formatting. Specifically, we slide a 5-second overlap on both sides of each 30-second training chunk, creating a 40-second "effective semantic window." Predictions are still made for the central 30 seconds, but the additional context improves semantic understanding. To implement this, we modified Whisper’s encoder to handle 40-second input sequences and fine-tuned the model on a subset of the VoxPopuli dataset \cite{wang-etal-2021-voxpopuli}.

Our approach reshapes the input structure during training. The audio is divided into 40-second chunks, where the middle 30 seconds are used for predictions, and the remaining 10 seconds (5 seconds on each side) provide additional context. Token prediction cross-entropy loss is calculated only for the central 30 seconds to ensure focus on the target window. This design enables the model to leverage a broader semantic context without affecting inference efficiency. Additionally, we handle entities that span chunk boundaries by assigning them entirely to the right-hand chunk, teaching the model to delay predictions when an entity does not fully fit within the central window. Figure \ref{fig:overall-windows} in the Appendix illustrates our windowing strategy.

During data preparation, we enriched transcripts by embedding entity labels (e.g., <LOC>United States</LOC>), allowing the model to learn both entity recognition and type-specific formatting. This step helps address the dual challenge of identifying entities and applying the correct formatting. For instance, numerical entities require precise formatting (e.g., phone numbers with punctuation), while long entities crossing chunk boundaries must be treated as single units.

To evaluate our approach, we used the Spoken Wikipedia dataset \cite{spokenwiki}, which provides 395 hours of professionally transcribed English audio. NER was evaluated using the SeqEval metric (precision, recall, F1-score, and accuracy) \cite{seqeval} . Entity formatting success was assessed for numerical entities (e.g., phone numbers, percentages) using Character Error Rate (CER) and for textual entities (e.g., names, locations) using Jaro-Winkler distance \cite{jaro1989,winkler1990}.

Our key contributions are summarized as follows:
\begin{itemize}
\item Extended Context Windows: We propose a novel training strategy using overlapping windows, enhancing the semantic context available to the ASR model. We analyze the impact of varying window lengths and incorporating future context on error metrics.
\item  Entity Formatting Task: We introduce entity formatting as a combined task of entity recognition and type-based formatting, with tailored evaluation metrics.
\item  Boundary Handling for Long Entities: We address the challenge of entities spanning chunk boundaries by teaching the model to delay predictions until the full context is available.
\item Evaluation by Entity Types: We evaluate numerical and non-numerical entities separately, using distinct metrics tailored to their characteristics.
\end{itemize}

Our results demonstrate significant improvements across all metrics, underscoring the value of extended semantic windows for ASR performance. By integrating semantic knowledge, our method establishes a new direction for context-aware transcription and entity recognition in long-form speech.

\begin{figure*}
    \centering
    \subfloat[All windows align smoothly. Both entities in the left and middle windows remain within boundaries, fitting perfectly.]{%
        \includegraphics[width=\textwidth]{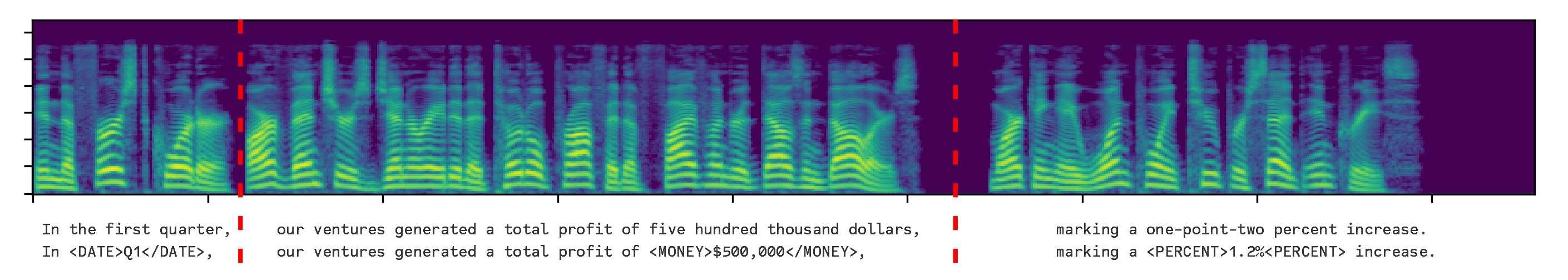} 
        \label{fig:subfigure1}
    }\\[0.3cm] 

    \subfloat[A numerical entity crosses from the middle to the right window. Since part of the entity crosses the boundary, we shift it entirely to the right. The same process can occur between the left and middle windows.]{%
        \includegraphics[width=\textwidth]{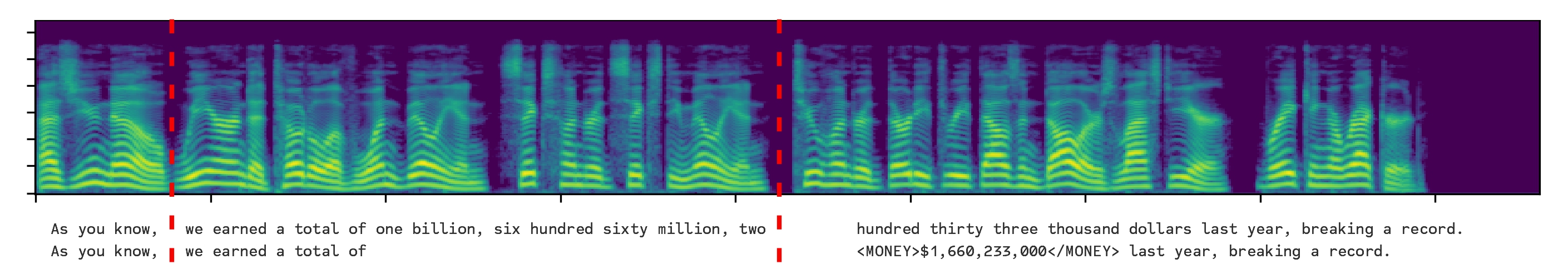} 
        \label{fig:subfigure3}
    }\\[0.3cm]

    \subfloat[An extreme case occurs here: two entities span across two boundaries. The entity on the left is shifted to the middle, and the middle entity is shifted to the right.]{%
        \includegraphics[width=\textwidth]{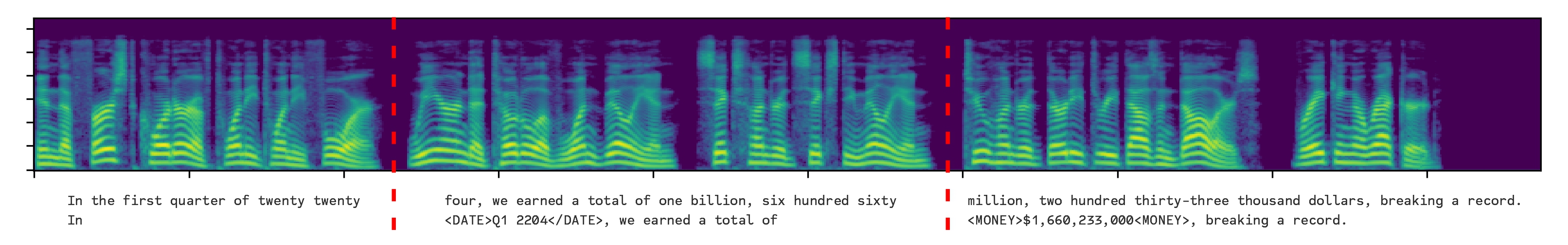} 
        \label{fig:subfigure4}
    }\\[0.3cm]

    \subfloat[No entities are shifted in this case. Entities are only adjusted when splitting them across boundaries is not feasible. Here, the textual entity remains in its normal setup.]{%
        \includegraphics[width=\textwidth]{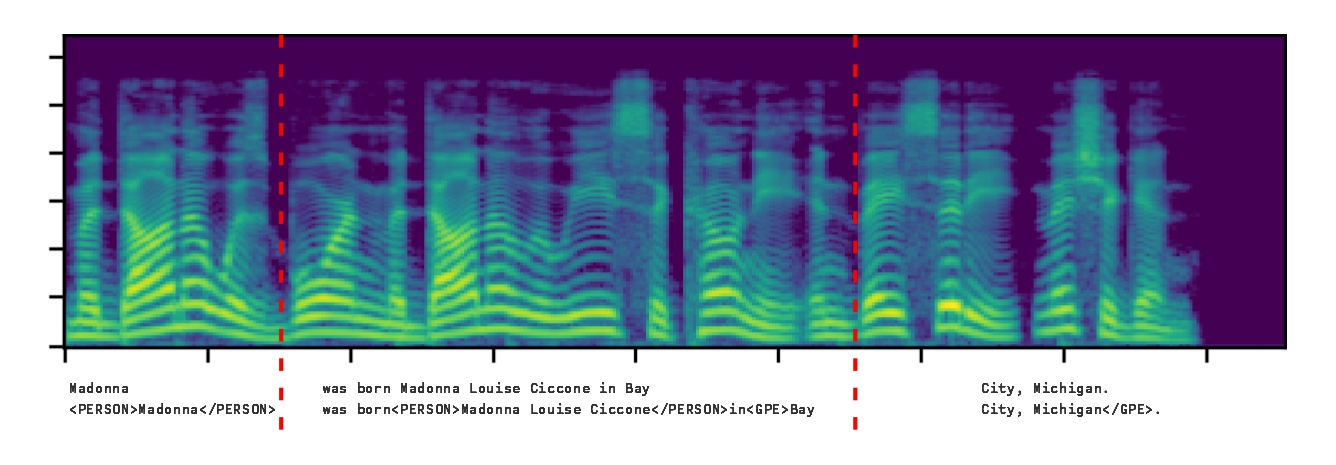} 
        \label{fig:subfigure5}
    }

    \caption{Examples illustrating our windowing methodology.}
    \label{fig:overall-windows}
\end{figure*}

\section{Background and Related Work}
\paragraph{Whisper} Whisper is a state-of-the-art sequence-to-sequence (seq2seq) ASR model developed by OpenAI. It is trained on 680,000 hours of multilingual and multitask supervised data collected from the web, making it one of the largest ASR models in terms of training data. Unlike many ASR models, Whisper is trained on unnormalized text, enabling it to handle diverse transcription styles and generate well-formatted text, including punctuation, casing, and numerical formatting. The model follows an encoder-decoder architecture, where the encoder processes input audio features, and the decoder produces the corresponding text transcription.

\paragraph{End-to-End NER in ASR} Earlier approaches to Named Entity Recognition (NER) in ASR relied on a two-step pipeline (ASR followed by NER). However, these approaches often suffered from format mismatches and transcription errors, including character deletions, insertions, and substitutions within entity spans \cite{szymanski-etal-2023-arent}.

Recent research has shifted towards end-to-end (E2E) approaches to mitigate these challenges. For example, Yadav et al. \cite{yadav2020endtoendnamedentityrecognition} incorporated bracketed entities during data preparation (e.g., [PERSON Nikola Tesla]) and trained a BiLSTM model with Connectionist Temporal Classification (CTC) loss. This method achieved improved F1 scores on the datasets LibriSpeech \cite{7178964}, Tedlium \cite{rousseau-etal-2012-ted} and Common Voice \cite{ardila2020commonvoicemassivelymultilingualspeech}. Their experiments focused on three entity types: person, location, and organization. Entities were extracted using the Flair NER tagger.

Similarly, Gaido et al. \cite{Gaido_2023} embedded entity tags directly into transcripts (e.g., <LOC>New York</LOC>), an approach comparable to the one used in this study. Although their work addressed speech translation rather than ASR, their methodology involved using spaCy NER \cite{spacy2} for transcript annotation, resulting in 18 entity types and 36 tags (including opening and closing tags).
\paragraph{Whisper and NER}Whisper's ability to generate formatted and punctuated text has made it a promising tool for NER tasks in ASR.

A recent study by Li et al. \cite{li2024usinglargelanguagemodel} combined Whisper with a large language model (LLM) using a fine-tuned adapter. Their experiments, conducted on the AISHELL-NER \cite{chen2022aishellnernamedentityrecognition} dataset, demonstrated significant improvements in NER performance and long-form ASR, including a 19.7\% relative reduction in character error rate (CER). The study focused on three entity types—person, location, and organization—annotated within the dataset. Additionally, entity tags were embedded into the transcripts to facilitate entity recognition and classification.

\section{Dataset}
\begin{figure}[ht]
    \centering
    \includegraphics[width=\columnwidth]{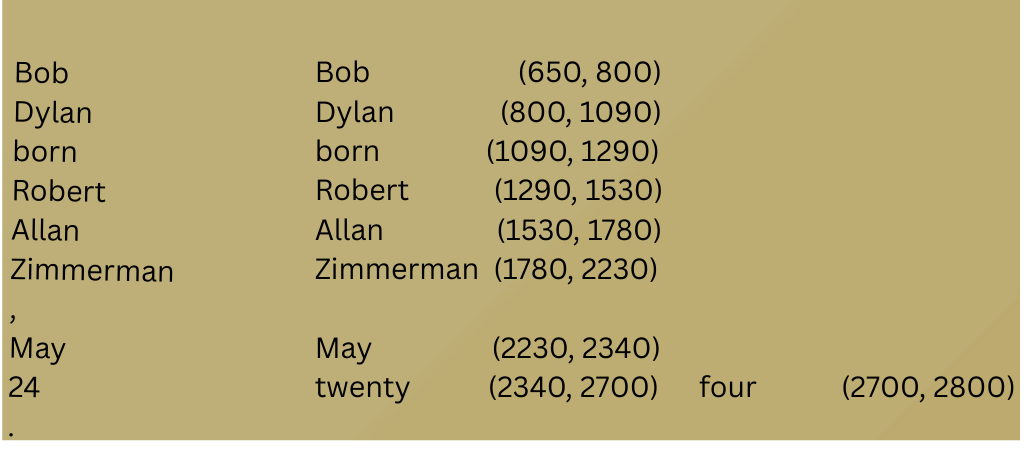}
     \caption{An example from the dataset, where each word is accompanied by its pronunciation, start time, and end time. This detailed alignment facilitates efficient chunking of the audio. Additionally, multi-word entities are annotated with word-level information for each constituent word.}
\label{fig:example-instance}
\end{figure}

This study utilizes the Spoken Wikipedia dataset, chosen for its long-form audio, rich entity vocabulary, and diverse entity types. The dataset comprises 1,339 English Wikipedia articles (395 hours of audio) and provides aligned audio and text in both normalized and formatted forms. It includes punctuation as separate tokens and word-level timestamps in XML format. An example instance from the dataset is illustrated in Figure \ref{fig:example-instance}.

On average, each instance contains 18 minutes of audio and 2,613 words, which aligns with the typical structure of Wikipedia articles. This characteristic provides substantial contextual information, making the dataset particularly suitable for semantic analysis.

We partitioned the dataset into 1,000 training instances (333 hours) and 241 test instances (62 hours). The corpus contains 3.5M words, including 472K entities spanning 828K words, accounting for 23\% of the total word count. Common entity types include person (103K), date (71K), location (60K), organization (59K), and cardinal number (42K). Numerical entity types collectively account for approximately 80K entities, representing 16\% of all entity types.

\begin{figure}[t]
  \centering
  \includegraphics[width=0.5\textwidth]{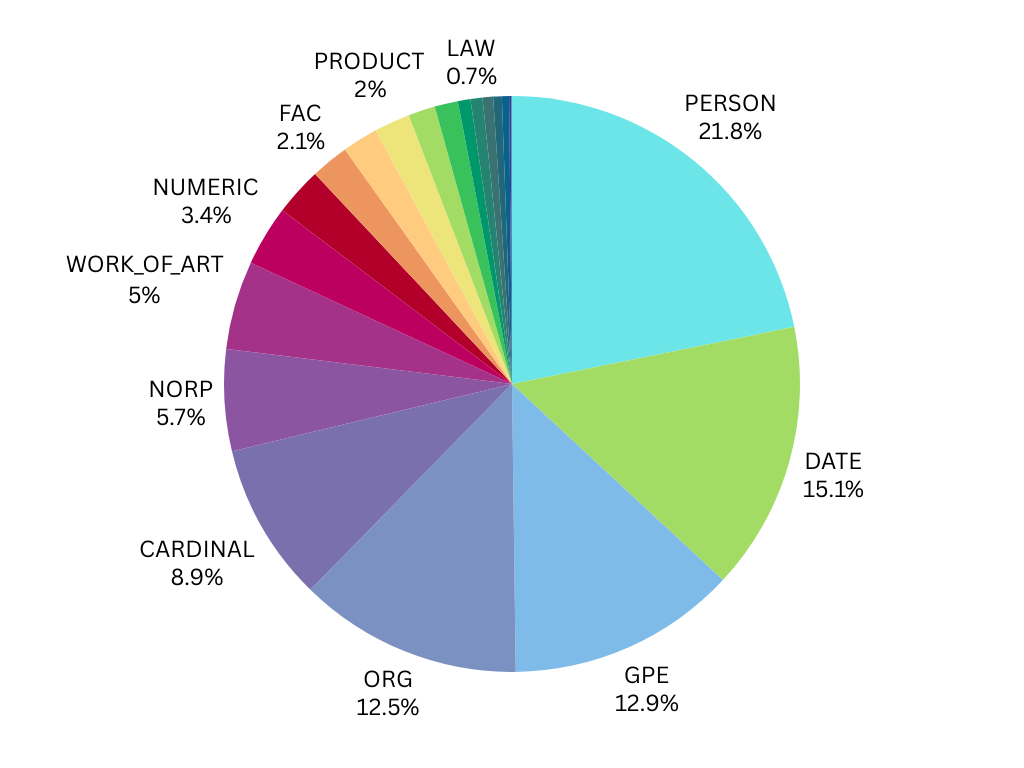}
  \caption{Distribution of NER labels in the dataset.}
  \label{fig:label-counts}
\end{figure}

Figure \ref{fig:label-counts} illustrates the distribution of entities across the dataset. A total of 22 NER labels were used: 18 derived from spaCy NER and 4 additional custom labels (URL, email, phone number, and generic numeric type). Entities were tagged using spaCy NER, while custom types were annotated using a regex-based extractor.

The dataset is particularly rich in numerical entity formats, which include examples such as \texttt{15,981.21}, \texttt{3,000}, \texttt{9.2\%}, \texttt{2.1 kg}, and \texttt{555-1142}, as well as email addresses and URLs. These types of entities are uncommon in standard speech datasets, making this corpus uniquely suited for tasks involving numerical and specialized entity recognition.

\section{Methodology}
We trained two models for this study. The first model, referred to as Whisper-puncted, is a baseline Whisper-Medium model trained on tagged transcripts with 30-second chunks. This model essentially represents a Whisper model that has learned entity information and serves as a baseline for comparison. The second model, called Whisper-puncted-windowed, is our proposed model. It is trained on 40-second chunks, comprising a 5-second left window, a 30-second middle segment, and a 5-second right window, with special handling for entities at chunk boundaries. Both models involved extensive data preparation and training processes, which are described in detail below.

The implementation and inference for both models were based on the Whisper-Medium model available from Hugging Face \cite{wolf2020huggingfacestransformersstateoftheartnatural}, utilizing Hugging Face's inference code.

\subsection{Data preparation}
The preparation of training data consisted of three phases:
\begin{itemize}
    \item Parsing XML input: The input XML was parsed to extract token-level indices and timestamps.
    \item  Chunking audio and text: The audio-text pairs were segmented into chunks of appropriate lengths.
    \item Reverse normalization: Text was reverse-normalized, and window boundaries were formed. 
\end{itemize}

\paragraph{Parsing XML} The first phase involved parsing the XML files to extract token indices and timestamps. Each token's position in the sentence was recorded for use in the reverse normalization process. The availability of token timestamps in the dataset made the chunking process straightforward.

\paragraph{Chunking} For Whisper-puncted, we segmented the audio into 30-second chunks with a 5-second stride. For Whisper-puncted-windowed, we used 40-second chunks, consisting of a 30-second middle segment and 5-second windows on the left and right. During this process, the left-mid-right windowed text and token indices were also collected. On average, each windowed chunk contained 58 words, while a standard chunk contained 42 words.

\paragraph{Reverse-normalization} The third phase involved implementing a reverse-normalizer, which combined a Named Entity Recognition (NER) model and a tag embedder. This process aimed to identify entity spans and embed the corresponding NER labels into the text.

The reverse-normalizer processed the entire transcript file rather than individual chunks to provide better context for the NER model. Entities were labeled with their start and end indices, which aligned with the token indices collected during the parsing phase. This allowed us to produce reverse-normalized chunk texts with precise entity annotations.

\begin{figure*}[ht]
    \centering
    \includegraphics[width=\textwidth]{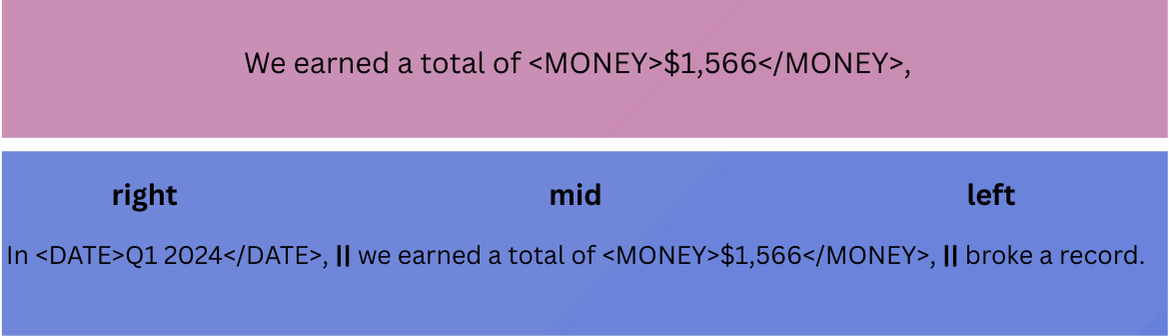}
\caption{Two examples of reverse normalization. The first example illustrates a reverse-normalized chunk text for the Whisper-puncted model. The second example demonstrates a reverse-normalized chunk text divided into left, middle, and right windows for the Whisper-puncted-windowed model.}
\label{fig:example-reverse-normalized}
\end{figure*}

For Whisper-puncted, the reverse-normalized chunk text was directly used. Special care was taken to ensure that single-token entities were not split at chunk boundaries, as this was not meaningful given the formatted text structure. Figure \ref{fig:example-reverse-normalized} illustrates an example reverse-normalized chunk.

For Whisper-puncted-windowed, the reverse-normalized chunk text was also used, with additional handling for entities that spanned chunk boundaries. The left, middle, and right boundary indices, as well as token indices from the chunker, were utilized to adjust entity spans. Entities extending across left-mid or mid-right boundaries were moved entirely into the next window. Entities split across the start of the left window or the end of the right window were not allowed during chunking. Examples of windowed chunks are shown in Figure \ref{fig:example-reverse-normalized}.

\begin{figure*}[ht]
    \centering
     \includegraphics[width=\textwidth]{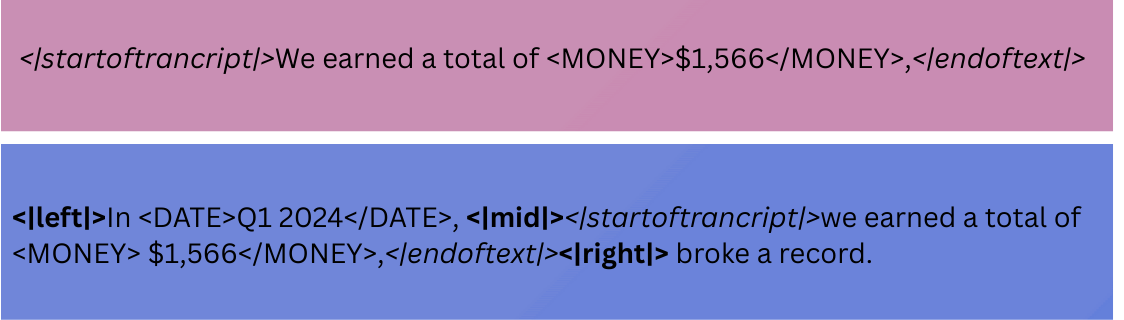}
\caption{Special token placement for a standard chunk and a windowed chunk..}
\label{fig:example-tokenized}
\end{figure*}

\subsection{Enhancing Whisper tokenizer} For both Whisper-puncted and Whisper-puncted-windowed, we introduced Named Entity Recognition (NER) opening and closing tags (e.g., <MONEY> and </MONEY>), adding a total of 44 tags. These tokens were incorporated as standard tokens to ensure that the model outputs them accurately during transcription.

For Whisper-puncted-windowed, we added three additional tokens: \texttt{<|left|>}, \texttt{<|mid|>}, and \texttt{<|right|>}. These were defined as special tokens, following Whisper's convention for special tokens.

One additional detail regarding the windowing approach is that the special tokens \texttt{<|startoftranscript|>} and \texttt{<|endoftext|>} were placed inside the mid window. This placement was intended to signal the model to generate text exclusively within this section. Figure \ref{fig:example-tokenized} illustrates the placement of these special tokens within the chunk text.

\subsection{Extending Whisper input length} For Whisper-puncted-segmented, we extended the input length to accommodate 40-second audio segments, as opposed to the default 30-second chunks. To achieve this, we reshaped the input embedding layer by adding additional rows. The new rows were initialized using the Glorot-uniform initialization method, ensuring the mean and variance were consistent with the original embeddings.

The decision to modify the network, rather than directly passing 40-second audio segments, was necessary due to Whisper's pre-trained architecture. Whisper's encoder is designed with a fixed number of positional embeddings that correspond to the default maximum input length of 30 seconds. Passing 40-second audio directly would result in the model either truncating the input or being unable to process the additional audio since the necessary positional embeddings would be missing. By extending the embedding layer, we ensured that the model could properly process and learn from the additional audio length without disrupting its internal structure or performance.

The newly added embeddings were trained using 100 hours of the VoxPopuli dataset. VoxPopuli provides fully punctuated sentences, though it does not include formatting for numbers. This limitation was acceptable for our purposes, as our primary goal was to train the new embeddings. Additionally, Whisper has already been extensively trained, mitigating the need for further extensive training.

Since VoxPopuli does not inherently include 40-second audio files, we concatenated shorter audio files from the dataset to create 40-second chunks for training. To ensure a natural flow in the concatenated segments, we inserted brief silences (typically 200–500 milliseconds) between the files, simulating pauses between consecutive utterances. This approach maintained the integrity of the dataset while making it compatible with the extended input length of the Whisper encoder.

Before concatenation, the VoxPopuli dataset consisted of shorter files, typically 5–15 seconds in length, each representing a single utterance or short sequence of utterances. After concatenation, the dataset was transformed into 40-second chunks, resulting in longer and contextually richer audio files while maintaining the natural acoustic characteristics of the original dataset.

\subsection{Learning the window boundaries}

\subsubsection{Problem setup}
Let the input audio sequence be represented as:
\begin{equation}
X = \{x_1, x_2, \dots, x_T\}    
\end{equation}
where \(T\) is the total number of audio frames. The target transcript is represented as:
\begin{equation}
Y = \{y_1, y_2, \dots, y_N\}    
\end{equation}
where \(N\) is the total number of tokens in the target transcript. The target transcript includes special tokens \( \langle \text{left} \rangle \), \( \langle \text{mid} \rangle \), and \( \langle \text{right} \rangle \), which define the boundaries of the windows.
The Whisper model maps the input audio \(X\) to hidden states:
\begin{equation}
H = \{h_1, h_2, \dots, h_M\}    
\end{equation}
through an encoder, and the decoder predicts logits:
\begin{equation}
L = \{l_1, l_2, \dots, l_N\}    
\end{equation}
where \(l_i \in \mathbb{R}^{|V|}\) is a probability distribution over the vocabulary \(V\) at time step \(i\).

\subsubsection{Label Smoothing}
Label smoothing regularizes the model by relaxing the one-hot target distribution. Let \(\bm{p}_i\) denote the smoothed target distribution for token \(y_i\), and \(\bm{q}_i\) be the predicted probability distribution obtained by applying softmax to the logits \(l_i\). 
The smoothed target distribution \(\bm{p}_i\) is defined as:

\begin{equation}
\bm{p}_i = (1 - \epsilon) \cdot \text{one\_hot}(y_i) + \frac{\epsilon}{|V|} \cdot \bm{1}    
\end{equation}

where $\epsilon$ is the label smoothing factor, \(\text{one\_hot}(y_i)\) is the one-hot encoding of the ground truth token \(y_i\),  \(|V|\) is the size of the vocabulary and \(\bm{1}\) is a uniform distribution over the vocabulary.

The cross-entropy loss with label smoothing is then given by:

\begin{equation}
\text{CE}(\bm{p}_i, \bm{q}_i) = - \sum_{v \in V} \bm{p}_i(v) \log \bm{q}_i(v)    
\end{equation}

where \(\bm{q}_i(v)\) is the predicted probability of token \(v\).

\subsubsection{Masking tokens outside the mid window}
To ensure the model focuses only on the tokens within the \( \langle \text{mid} \rangle \) window, we define a mask \(\bm{m} \in \{0, 1\}^{N}\) over the token sequence \(Y\). 

Let \(t_{\text{mid}}\) and \(t_{\text{right}}\) be the positions of the \( \langle \text{mid} \rangle \) and \( \langle \text{right} \rangle \) tokens, respectively. The mask \(\bm{m}\) is defined as:

\begin{equation}
\bm{m}_i =
\begin{cases} 
1 & \text{if } t_{\text{mid}} \leq i < t_{\text{right}} \\
0 & \text{otherwise}
\end{cases}
\end{equation}

The mask ensures that only tokens between \( \langle \text{mid} \rangle \) and \( \langle \text{right} \rangle \) are included in the loss computation.

\subsubsection{Final Cross-Entropy loss}
The final loss combines label smoothing and masking. Let \(\text{CE} = \{\text{CE}(\bm{p}_i, \bm{q}_i)\}_{i=1}^N\) denote the cross-entropy loss for each token. The masked loss is computed as:

\begin{equation}
\text{Loss} = \frac{\sum_{i=1}^N \bm{m}_i \cdot \text{CE}_i}{\sum_{i=1}^N \bm{m}_i}
\end{equation}

where:
\begin{itemize}
\item \(\bm{m}_i \cdot \text{CE}_i\) includes the loss only for the tokens between \( \langle \text{mid} \rangle \) and \( \langle \text{right} \rangle \),
\item \(\sum_{i=1}^N \bm{m}_i\) normalizes the loss by the number of valid tokens, avoiding division by zero.
\end{itemize}

\subsection{Inference} 
For long audio, the Whisper inference pipeline uses striding to maintain context across segments. The audio is divided into fixed-length segments with a stride value, effectively applying windowing. Each segment is transcribed sequentially. To merge consecutive segments, the tokens in the overlapping stride must be calculated. Consequently, the inference pipeline requires the token IDs and timestamps from the previous segments to compute the intersection introduced by striding during runtime. Typically, Whisper generates timestamp tokens, which can be leveraged without printing special tokens during text generation.

Our base model, Whisper-puncted, is an ordinary Whisper model fine-tuned on entity-tagged data. As such, we used the HuggingFace audio pipeline directly. However, in our setup, all added token IDs were greater than the timestamp token IDs, which caused the model to fail in emitting any timestamp tokens—an essential component for long-audio inference. To address this, during the training of Whisper-puncted, we added timestamp tokens to the start, middle, and end of the chunk text.

The inference logic for Whisper-puncted-windowed differs significantly from the standard Whisper setup. During runtime, the model processes 35-second audio segments (except for the first segment) in overlapping windows: \textit{left + mid}. The first segment, lacking a left context, is limited to 30 seconds. After the initial transcription, a 5-second stride is applied, and subsequent audio and text are fed in 35-second chunks. To retrieve the final 5 seconds from the preceding segment, timestamp tokens are necessary.

We prompt the model with the left window transcript and special boundary tokens (e.g., \texttt{<|left|>}In this year,\texttt{<|mid|>}), then feed 35 seconds of audio containing both the \textit{right window's audio} and the mid window's audio. This ensures the model has sufficient semantic context for predictions while maintaining alignment with the windowed approach.

Unlike Whisper-puncted, merging text across segments is not required for Whisper-puncted-windowed. The 5-second stride occurs between the \texttt{<|left|>} and \texttt{<|mid|>} windows, where the model is explicitly trained not to emit any text. Recall that the model is designed to generate text exclusively for the mid window (the central 30 seconds). This design ensures clean segment boundaries and avoids generating redundant or incorrect predictions for overlapping regions.

\subsection{Evaluation metrics}
We utilize SeqEval for evaluating Named Entity Recognition (NER) performance, calculating the F1 score and accuracy for capitalization and punctuation, and assessing transcription quality using Character Error Rate (CER) for numerical entities and Jaro-Winkler distance for non-numerical entities. These metrics enable a comprehensive evaluation of transcription quality across both syntax and semantics.

Entity formatting is assessed based on the type of entity. For numerical entities, such as monetary values, percentages, or drug doses, we use CER to ensure strict digit-level precision. For example, the values \texttt{\$15,000,000} and \texttt{\$15,000} differ significantly in meaning, as they represent entirely separate amounts. Similarly, percentages like \texttt{9.5\%} and \texttt{119.5\%}, or drug doses such as \texttt{0.1mg} and \texttt{11.1mg}, convey vastly different information. Accurate numerical formatting is critical to preserve the semantic integrity of the transcription.

In contrast, textual entities are evaluated using Jaro-Winkler distance, which measures string similarity and accommodates minor variations. For instance, transcribing \texttt{Bryan Adams} as \texttt{Bryan Adam} is not considered a significant error compared to the strict requirements for numerical entities. This approach ensures the evaluation is appropriately tailored to the nature of each entity type.

\section{Experimental Setup}
All experiments were conducted using the HuggingFace Trainer code with Flax, and training was performed on a V4-8 TPU. The code is publicly available on our GitHub repository\footnote{\url{https://github.com/DuyguA/TSD2025-Mind-the-Gap}}, and all fine-tuned models can be accessed on our HuggingFace repository\footnote{\url{https://huggingface.co/BayanDuygu/whisper-puncted-timed}}.

\paragraph{Custom dataset and feature caching} The dataset was chunked into segments of 30 seconds and 40 seconds for the two respective models. Chunk text and timestamp information were stored in JSON format. Prior to training, audio features were extracted for each chunk and saved in NumPy .npz format on the local file system. A custom PyTorch dataset was implemented to provide chunks by index and fetch the corresponding audio features from the file system.

\paragraph{Whisper-puncted} The Whisper-puncted model was trained for 100 epochs. The best-performing checkpoint was recorded at the 80th epoch. Before the 20th epoch, the model struggled to fully learn the tags, and prior to the 50th epoch, it exhibited frequent hallucinations around entities. 

\paragraph{Whisper-puncted-windowed} This model was trained for 150 epochs. An additional 50 epochs were required to effectively learn the window boundaries.

For both models, a batch size of $32$ and a learning rate of $1e-4$ were used. The warmup steps were set to 10\% of the total training steps. The AdamW optimizer \cite{loshchilov2019decoupledweightdecayregularization} was employed with a weight decay of $0.01$. All training code was implemented in Flax.

\begin{table*}[th]
  \centering
  \setlength{\tabcolsep}{6pt}
  \begin{tabular}{lcccc|ccc}
    \toprule
    \textbf{Label} & \textbf{Ref. Count} & \multicolumn{3}{c}{\textbf{Whisper-puncted}} & \multicolumn{3}{c}{\textbf{Whisper-puncted-windowed}} \\ 
    \cmidrule(lr){3-5} \cmidrule(lr){6-8}
                   &                     & \textbf{F1} & \textbf{Prec.} & \textbf{Recall} 
                   & \textbf{F1} & \textbf{Prec.} & \textbf{Recall} \\
    \midrule
    PERSON         & 4212  & 0.50 & 0.40 & 0.67   & 0.65  & 0.50  & 0.77 \\
    DATE           & 3383  & 0.65 & 0.55 & 0.66   & 0.65  & 0.65  & 0.66 \\
    GPE            & 2462  & 0.61 & 0.58 & 0.65   & 0.71  & 0.70  & 0.70 \\
    CARDINAL       & 2212  & 0.95 & 0.96 & 0.94  & 0.98 & 0.98 & 0.98 \\
    QUANTITY       & 351  & 0.91 & 0.88  & 0.94   & 0.96  & 0.96 & 0.95 \\
    MONEY          & 109  & 0.92 & 0.93 & 0.91   & 0.96  & 0.96  & 0.96 \\
    \bottomrule
  \end{tabular}
     \caption{NER statistics of the Whisper-puncted and Whisper-puncted-windowed models on the most common textual and numeric entity types.}
  \label{tab:ner-base}
\end{table*}

\begin{table*}[th]
  \setlength{\tabcolsep}{6pt} 
  \centering
\begin{tabular}{lcc|lcc}
    \toprule
    \cmidrule(lr){1-3} \cmidrule(lr){4-6}
    \textbf{Label J-W} & \textbf{W-punct} & \textbf{W-pwindow} & \textbf{Label CER} & \textbf{W-punct} & \textbf{W-pwindow} \\
    \midrule
    PERSON         & 0.75    & 0.75    & CARDINAL       & 0.25    & 0.12    \\
    DATE           & 0.95    & 0.96    & NUMERIC        & 0.22    & 0.12    \\
    GPE            & 0.60    & 0.66    & TIME           & 0.41    & 0.18    \\
    NORP           & 0.96    & 0.96    & QUANTITY       & 0.22    & 0.08    \\ 
    ORG            & 0.71    & 0.71    & MONEY          & 0.42    & 0.11    \\
    EVENT          & 0.70    & 0.70    & PERCENT        & 0.32    & 0.12    \\
    LOC            & 0.67    & 0.68    & URL            & 0.57    & 0.32    \\
    ORD            & 0.87    & 0.94    & EMAIL          & 0.58    & 0.28    \\
    LAW            & 0.76    & 0.76    & PHONE\_NUM     & 0.35    & 0.15    \\
    FAC            & 0.57    & 0.57    &                &         &         \\
    PROD           & 0.69    & 0.69    &                &         &         \\
    LANG           & 0.90    & 0.90    &                &         &         \\
    W\_OF\_ART     & 0.47    & 0.47    &                &         &         \\
    \bottomrule
\end{tabular}
  \caption{Per-label statistics by Jaro-Winkler and CER scores for Whisper-puncted and Whisper-puncted-windowed models.}
  \label{tab:combined_stats}
\end{table*}

\section{Results and Discussion}

\subsection{Baseline}
The Whisper-Medium model achieved a Word Error Rate (WER) of 38\% on The Spoken Wiki dataset, which is reasonable given the complexity of the transcripts. Our fine-tuned version, Whisper-puncted, reduced the WER to 26\%. For the WER calculations, punctuation, named entity recognition (NER) tags, and text case were normalized by removing punctuation, converting text to lowercase, and ignoring NER tags.

Table \ref{tab:ner-base} presents the sequence evaluation (seqeval) results. Numerical entities, such as dates, performed well due to their clear wording and minimal context dependence. High match scores were observed for dates, NORP (e.g., nationalities, religious groups), and ordinal numbers. However, geographical locations, product names, and works of art performed poorly. Surprisingly, person names achieved reasonable accuracy, likely due to Wikipedia's bias toward celebrity names. Numerical entities were more accurately tagged due to their distinct formatting, which likely contributed to these results.

\begin{figure*}[ht]
    \centering
    \includegraphics[width=\textwidth]{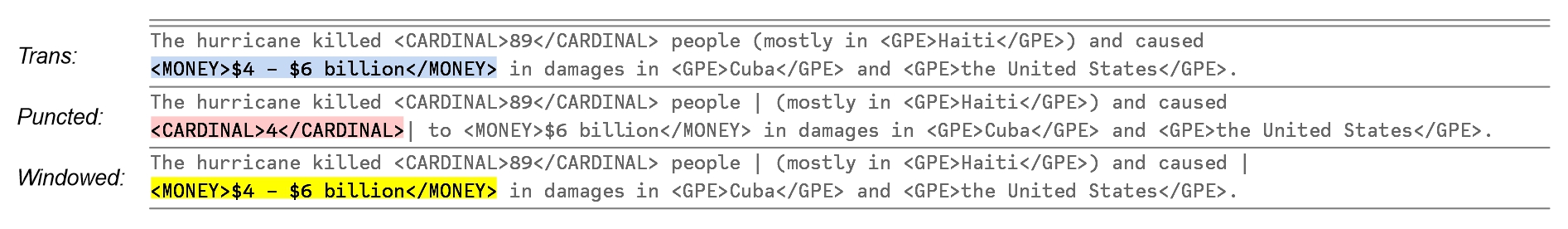} 
    \caption{An example of a numerical entity transcription error caused by spreading across chunk boundaries. The first word of the entity, \texttt{\$4-\$6 billion}—"four"—was predicted within the first chunk, while the remaining words of the entity were predicted in the second chunk. Having only the word "four" in the first chunk did not provide the model with sufficient context to recognize it as a monetary amount, leading to the incorrect prediction of a cardinal tag. The presence of "dollars" in the second chunk provided the necessary cues for identifying it as a monetary entity. However, the resulting transcription is unusable and contributes to an increased WER.}
    \label{fig:spread-entities} 
\end{figure*}

Table \ref{tab:combined_stats} summarizes the entity formatting success for each label. The Character Error Rate (CER) for numerical entities was higher than expected, often due to incorrect punctuation in numbers (e.g., "\$1.2.211"). Despite training, Whisper struggled with non-sentence punctuation, particularly within entities. Another source of errors was long numerical entities spread across chunk boundaries, which we aimed to address through our windowing scheme (Figure \ref{fig:spread-entities}).

\subsection{Windowing Approach}
Our Whisper-puncted-windowed model improved both NER success and formatting success for textual and numerical entities. Table \ref{tab:ner-base} shows improvements in NER performance, as feeding the model more context allowed for more accurate entity extraction.

For entity formatting success, numerical entities showed significant improvement, while textual entities saw marginal gains. We attribute this to textual entities having more contextual clues in the text. Additionally, punctuation errors within entities were reduced, as the model learned to handle entity tags and the content between them more effectively. Compacting numerical entities at chunk boundaries also contributed significantly to the improvements. Table \ref{tab:combined_stats} shows the results.

\section{Conclusion}
In conclusion, our work demonstrates the challenges and solutions for improving transcription quality in long-form audio tasks, particularly for numerical and textual entities. By leveraging fine-tuned models such as Whisper-puncted and Whisper-puncted-windowed, alongside custom dataset preprocessing and feature caching techniques, we achieved significant advancements in entity recognition and transcription accuracy. While both models required careful training setups to address issues like hallucination and boundary learning, the results highlight the potential of tailored approaches for handling complex audio data. Our publicly available code and models pave the way for further research and development in this domain.

\section{Acknowledgements}
We thank the Google Cloud TPU Research Program for providing the computational resources that made this work possible. Their support was instrumental in enabling the training and evaluation of our models on TPU infrastructure. Additionally, we acknowledge the use of OpenAI's language model GPT-4 for assisting with text formatting and improving the clarity of the manuscript.

\begin{raggedright}
\bibliography{custom}
\end{raggedright}

\end{document}